 \documentclass[pmlr,twocolumn]{jmlr} 

\usepackage{booktabs}
\usepackage[load-configurations=version-1]{siunitx} 

\theorembodyfont{\upshape}
\theoremheaderfont{\scshape}
\theorempostheader{:}
\theoremsep{\newline}

\jmlrworkshop{Machine Learning for Health (ML4H) 2020} 

\title[Understanding Clinical Context of Medication Events]{Toward Understanding Clinical Context of Medication Change Events in Clinical Narratives}

  \author{
   \Name{Diwakar Mahajan} \Email{dmahaja@us.ibm.com}\\
   \Name{Jennifer J. Liang} \Email{jjliang@us.ibm.com}\\
   \Name{Ching-Huei Tsou} \Email{ctsou@us.ibm.com}\\
   \addr T.J. Watson Research Center, Yorktown Heights, NY, USA
   }

\begin{document}

\maketitle

\begin{abstract}
Understanding medication events in clinical narratives is essential to achieving a complete picture of a patient's medication history. While prior research has explored classification of medication changes from clinical notes, studies to date have not considered the necessary clinical context needed for their use in real-world applications, such as medication timeline generation and medication reconciliation. In this paper, we present the Contextualized Medication Event Dataset (CMED), a dataset for capturing relevant context of medication changes documented in clinical notes, which was developed using a novel conceptual framework that organizes context for clinical events into various orthogonal dimensions. In this process, we define specific contextual aspects pertinent to medication change events, characterize the dataset, and report the results of preliminary experiments. CMED consists of 9,013 medication mentions annotated over 500 clinical notes, and will be released to the community as a shared task in 2021.
\end{abstract}

\section{Introduction}
\label{sec:intro}
An accurate medication history is foundational for providing quality medical care, allowing healthcare providers to better assess the appropriateness of current treatments, detect potential medication-related pathologies or symptoms, and direct future treatment options. Although providers usually rely on structured medication orders for such information, many medication events are documented only in unstructured clinical notes, which contain richer medication information but are difficult to search through.
As clinical documentation captures the longitudinal patient history including providers' reasoning behind medical decisions, extraction of medication changes alone without the necessary clinical context is insufficient for use in real-world applications, such as medication timelines or medication reconciliation.

Previous research in classifying medication changes in clinical notes includes work that target specific medications \citep{liu2011modeling,meystre2015heart,fan2016classification,fan2018using} and those that cover all medications \citep{sohn2010classification,lerner2020learning,pakhomov2002maximum}. However, these works, driven by a variety of use cases, have resulted in a mixed set of labels, some of which may not cover all types of changes (e.g. labels used by \citet{pakhomov2002maximum} - \emph{past, continuing, stop, start}, do not differentiate between increase and decrease), or do not cover all aspects of a medication event (e.g. labels used by \citet{sohn2010classification} - \emph{start, stop, increase, decrease, no-change}, cover all types of changes but do not provide temporal information).
Another body of work has investigated recognizing context or assertions for medical concepts such as problems and tests \citep{harkema2009context,uzuner20112010,mowery2013semantic,rumeng2017hybrid,albright2013towards}. 
Although these works identify aspects like certainty, none have been applied to medications. Further, no previous work has attempted to identify an event's actor, which is especially important for medications due to implications for patient adherence.
Thus, there is need for a more organized schema of label definitions that better contextualizes medication events.

To address this need, here we propose a conceptual framework to organize multi-dimensional context for medication events in clinical narratives, and present the resulting dataset - Contextualized Medication Event Dataset (CMED). CMED captures pertinent context for medication events along four orthogonal dimensions, and will be released to the community as a shared task in 2021. We present an overview of our annotation guidelines and challenges, and provide an initial baseline to assess the feasibility of this task.

\vspace{-4mm}
\section{Annotation Task and Data}
\label{sec:math}
For this study, we used clinical notes from the 2014 i2b2/UTHealth Natural Language Processing shared task \citep{kumar2015creation, stubbs2015automated, stubbs2015identifying}, which contains 2-5 notes per patient, 
to allow potential future work in reconciling medication events across different notes. 
From this corpus, 500 notes were randomly selected for annotation by a team of three annotators led by a physician, of which 120 notes were doubly-annotated and adjudicated to measure inter-annotator agreement (IAA) using Cohen's kappa. 

\vspace{-2mm}
\subsection{Annotation Guidelines}
\label{sec:guidelines}
\begin{table*}
\setlength{\tabcolsep}{5pt} 
\renewcommand{\arraystretch}{1} 
\caption{\label{table:examples} Sample annotations demonstrating how labels change depending on the context.}
\scriptsize
\centering
\begin{tabular}{|m{4.4cm}|>{\centering\arraybackslash}m{1.9cm}|>{\centering\arraybackslash}m{1.8cm}|>{\centering\arraybackslash}m{1.9cm}|>{\centering\arraybackslash}m{1.7cm}|>{\centering\arraybackslash}m{1.3cm}|}
\hline
\textbf{Text} & \textbf{Event} & \textbf{Action} & \textbf{Temporality} & \textbf{Certainty} & \textbf{Actor} \\
\hline
Pt currently on \emph{lisinopril} & NoDisposition & -- & -- & -- & --  \\ \hline
Plan: incr \emph{losartan} from 1 tab qd to bid. & Disposition & Increase & Present & Certain & Physician \\ \hline
If BP$<$100 hold off on taking \emph{hctz} & Disposition & Stop  & Future & Conditional & Physician   \\ \hline
In ED, given \emph{ativan} 1 mg IV x 1 & Disposition & UniqueDose & Past & Certain & Physician \\ \hline
She was experiencing a bad episode of dry cough so stopped taking \emph{lisinopril} & Disposition & Stop  & Past & Certain & Patient   \\ \hline
On \emph{Zocor}, pt wants to discuss switching to generic to save money & Disposition & OtherChange & Present & Hypothetical & Patient \\ \hline
\end{tabular}
\end{table*}
We define a medication event as any discussion about a medication change in a given patient. 
The annotation process is as follows. For each medication mention, 
the annotator first determines if a change is being discussed (Disposition) or not (NoDisposition); if more information is needed to determine this, the mention is labeled as Undetermined. Next, for identified Disposition events, the annotator labels the multi-dimensional context along four orthogonal dimensions:
\vspace{-2mm}
\begin{itemize}
\addtolength\itemsep{-3mm}
    \item \textbf{Action}: What is the change discussed? (Start, Stop, Increase, Decrease, OtherChange, UniqueDose, Unknown)
    \item \textbf{Temporality}: When is this change intended to occur? (Past, Present, Future, Unknown)
    \item \textbf{Certainty}: How likely is this change to have occurred / will occur? (Certain, Hypothetical, Conditional, Unknown)
    \item \textbf{Actor}: Who initiated the change? (Physician, Patient, Unknown)
\end{itemize}
\vspace{-2mm}
Note that a mention can have 0, 1, or more Disposition events. For example,  
``\emph{Started metformin at last visit but pt stopped after 1wk because of GI upset}'' has two events for \emph{metformin} (Start\textbar Past\textbar Certain\textbar Physician and Stop\textbar Past\textbar Certain\textbar Patient).
These dimensions were designed to capture a wide variety of contexts, such as references to past events (Temporality: Past), treatments being considered but not yet decided upon (Certainty: Hypothetical), or episodes of patient nonadherence (Actor: Patient). Some sample annotations are shown in Table \ref{table:examples}. To assist annotators, medication mentions were pre-annotated using a medication extraction model and corrected as necessary during annotation. Detailed label definitions are presented in \appendixref{apd:first}.
\vspace{-4mm}
\subsection{Annotation Challenges}

Knowledge of a medication's attributes can affect label assignment. 
For example, ``\emph{Plan: Z-Pak}'' could be labeled as Undetermined because it could either be restating an ongoing medication or indicating the start of a new medication. 
However, knowing that Z-Pak is a 5-day antibiotic course, one would more likely label this as a start action. 
Similarly, knowing whether a medication is over-the-counter or requires a prescription can affect whether the Actor is more likely to be Physician or Patient. 
To best reflect the reality of the event, annotators were instructed to annotate based on what they think is actually going on, even if the annotation depends on medical knowledge not in the text.

Ambiguous language in clinical text can lead to confusion in label assignment. For example, grammatical tense is not always indicative of when an event takes place. 
Since providers can copy-and-paste text from previous notes \citep{shoolin2013copypaste}, a sentence that appears to be Present based on tense (e.g. ``\emph{Start plavix}''), when in the Past Medical History (PMH) section, actually indicates a Past action. To address this, we asked annotators to consider the surrounding context when assigning labels, roughly defined as the sentence containing the medication mention +/- 1 sentence and the note section.
\vspace{-2mm}
\subsection{Data Statistics}

IAA on classifying whether a medication change is discussed (Disposition vs NoDisposition vs Undetermined) was 0.88 on 2,495 annotated medication mentions.
For the 367 instances of agreed Disposition events, IAA for each dimension was: Action (0.87), Temporality (0.94), Certainty (0.75), Actor (0.72). 
The lower IAA in Certainty was due to unclear language, 
e.g. ``\emph{We might suggest that she be started on Cisapride 10 mg qd}''.
Disagreements in Actor were mostly for past events, e.g. ``\emph{Stopped Detrol}'', where the subject is unspecified and was interpreted differently by different annotators.

As CMED will be used in a shared task in 2021, we split the final dataset 80/20 for train and test sets, and only report statistics on the train set in this paper. The train set consists of 7,230 annotated medication mentions over 400 notes. The specific label distribution is presented in Table \ref{table:experiments}.

\vspace{-5mm}
\section{Method and Experiments}
\label{section:experiments}
\begin{table*}
\scriptsize
\caption{Ablation results (F1-score) for 5-fold cross validation on train set with label distribution. + indicates the addition of each feature class to the base n-grams model.}
\label{table:experiments}
\setlength{\tabcolsep}{2.7pt} 
\renewcommand{\arraystretch}{1} 
\begin{tabular}{|c|ccc|ccc|cc|}
\hline
& \multicolumn{3}{c||}{\textbf{Event}} & \multicolumn{5}{c|}{\textbf{Action}} \\
\hline
\textbf{Experiment} & \textbf{Disp} & \textbf{NoDisp} & \multicolumn{1}{c||}{\textbf{Undet}} & \textbf{Start} & \textbf{Stop} & \multicolumn{1}{c}{\textbf{Increase}} & \textbf{Decrease} & \textbf{UniqueDose}
\\
\hline
n-grams & 0.64   & 0.86 & \multicolumn{1}{c||}{0.32} & 0.67 & 0.58 & \multicolumn{1}{c}{0.66} & 0.17 & 0.74 \\
\hline
+lexico-syntactic & 0.62 & 0.85 & \multicolumn{1}{c||}{0.30} & 0.66 & 0.58 & \multicolumn{1}{c}{\textbf{0.67}} & 0.15 & 0.74 \\
\hline
+window & \textbf{0.65} & 0.87 & \multicolumn{1}{c||}{0.33} & \textbf{0.72} & \textbf{0.62} & \multicolumn{1}{c}{\textbf{0.67}} & \textbf{0.19} & \textbf{0.77} \\
\hline
+dependency-parse & 0.64 & 0.87 & \multicolumn{1}{c||}{0.31} & 0.70 & 0.59 & \multicolumn{1}{c}{0.66} & 0.09 & 0.75 \\
\hline
+note-section & 0.62 & \textbf{0.89} & \multicolumn{1}{c||}{0.35} & 0.55 & 0.46 & \multicolumn{1}{c}{0.34} & 0.08 & 0.65 \\
\hline
+RxNorm &\textbf{0.65}& 0.88 & \multicolumn{1}{c||}{0.31} & 0.71 & 0.58 & \multicolumn{1}{c}{0.63} & 0.11 & 0.72 \\
\hline
ALL & \textbf{0.65} & 0.88 & \multicolumn{1}{c||}{\textbf{0.38}} & 0.68 & 0.58 & \multicolumn{1}{c}{0.62} & 0.10 & 0.75 \\
\hline
\textbf{Counts} & 1413 & 5260 & \multicolumn{1}{c||}{557} & 568 & 341 & \multicolumn{1}{c}{129} & 54 & 285 \\
\hline
\hline
& \multicolumn{3}{c|}{\textbf{Temporality}} & 
\multicolumn{3}{c|}{\textbf{Certainty}} & 
\multicolumn{2}{c|}{\textbf{Actor}} \\
\hline
\textbf{Experiment} & \textbf{Past} & \textbf{Present} & \textbf{Future} & \textbf{Certain} & \textbf{Hypothetical} & \textbf{Conditional} & \textbf{Physician} & \textbf{Patient}
\\
\hline
n-grams & 0.85 & \textbf{0.69} & 0.36 & 0.92  & 0.34 & 0.38 & \textbf{0.96} & 0.38\\
\hline
+lexico-syntactic & 0.82 & 0.65 &0.35  &0.91  & 0.40 & \textbf{0.39} & 0.95 & 0.37\\
\hline
+window & 0.85 & 0.67 &0.33  & \textbf{0.93}  & 0.40 & 0.40 & \textbf{0.96} & 0.44\\
\hline
+dependency-parse & 0.85 & 0.68 &0.36  &0.92  & 0.38 & 0.43 & \textbf{0.96} & 0.41\\
\hline
+note-section & \textbf{0.86} & \textbf{0.69} & 0.35  & 0.92  & \textbf{0.42} & 0.38 & 0.94 & \textbf{0.45}\\
\hline
+RxNorm & 0.85 & 0.68 &0.35  &0.92 & 0.32 & 0.38 & \textbf{0.96} & 0.36\\
\hline
ALL & 0.85 & 0.68  & \textbf{0.39} & 0.92  & 0.41 & 0.37 & \textbf{0.96} & 0.43\\
\hline
\textbf{Counts}& 745 & 494 & 145 & 1177 & 134 & 100 & 1278 & 107\\
\hline
\end{tabular}
\end{table*}

We split the overall task into 5 classification sub-tasks organized in a two-step process: (1) one sub-task to classify each medication mention into event type - Disposition, NoDisposition, Undetermined, and (2) four sub-tasks to classify each Disposition event along the four context dimensions. 
To explore the dataset and various dimensions, we elected to use a feature-based approach with a discriminative classification algorithm in our experiments (SVM with linear kernel). Our features were divided into the following classes: \emph{n-grams}, \emph{lexico-syntactic (POS tags \&  lemma)},
\emph{window-based}, \emph{dependency-parse},  \emph{note-section}, and \emph{RxNorm} features. To deal with skew in class distribution, we adjusted class weights to be inversely proportional to class frequencies.  OtherChange for Action dimension and Unknown for all dimensions were excluded due to the limited number of instances (\textless40). Table \ref{table:experiments} shows ablation results for the 5 sub-tasks employing 5-fold cross validation on the train set. Detailed explanation of the features and other experimental settings are in \appendixref{apd:second}.

For the first sub-task i.e. event classification,
\emph{note-section}, \emph{window} and \emph{RxNorm} features were most impactful as the location of the medication mention along with textual clues provide necessary information to classify a medication event. 
For the next four sub-tasks, 
different dimensions benefited from different features.
Action and Certainty dimensions were largely impacted by \emph{window} and \emph{lexico-syntactic} features as they are highly dependent upon textual clues. Temporality and Actor benefited from \emph{note-section}, as the location of the medication within the note can inform its chronology (e.g. History of Present Illness (HPI) vs PMH) and actor (e.g. patient-initiated changes are usually recorded in the HPI section). 
As different features affect dimensions differently, further exploration is needed towards dimension-specific feature selection.

\vspace{-9mm}
\section{Conclusion}
\vspace{-2mm}
We presented CMED, a dataset for capturing relevant context needed to understand medication changes in clinical narratives.
CMED not only identifies medication changes, but also contextualizes them along four dimensions (Action, Temporality, Certainty, Actor), providing a more well-rounded view of medication change events in clinical notes. Notably, this multi-dimensional approach for context extraction is extensible and can be adapted to other types of events, e.g. adverse drug events, procedures. 
To fully utilize the extracted contextual information, additional research is needed for temporal resolution (placing Past and Future events at a more precise point in time) and coreference resolution (resolving multiple references to the same event) which we plan to explore as part of our future work.
\bibliography{jmlr-sample}

\appendix
\newpage
\section{Annotation Label Definitions}\label{apd:first}

\begin{figure*}[]
\centering
\includegraphics[scale=0.48]{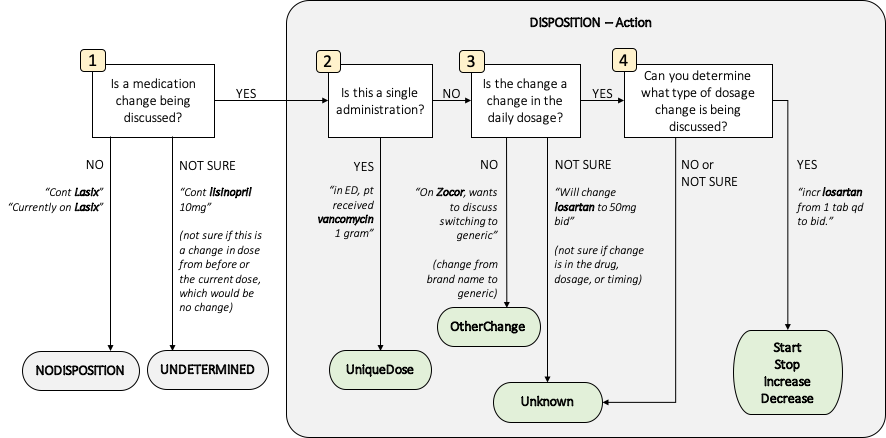}
\caption{Flowchart demonstrating stepwise thought process for annotating event type (Disposition, NoDisposition, Undetermined) for medication mentions, and action label for Disposition events.}
\label{fig:flowchart}
\end{figure*}

For each medication mention within a clinical note, annotators were asked to first determine whether a medication change is being considered or discussed.

\begin{itemize}
\addtolength\itemsep{-3mm}
    \item \textbf{NoDisposition}: (1) no action is being discussed, e.g. status statements (``\emph{doing well on 10mg lisinopril}''), or (2) continue statements with no dosage details (``\emph{continue lisinopril as prescribed}'')
    \item \textbf{Disposition}: presence of a medication change action being discussed
    \item \textbf{Undetermined}: unclear if a change is being discussed, for example, ``\emph{Plan: Lasix}'' -- unclear if just stating a medication patient is on (NoDisposition) or starting a new medication (Disposition), or ``\emph{Continue Lasix 10mg}'' -- not sure if this is a change in dose from before (Disposition) or the current dose (NoDisposition)
\end{itemize}

Next, for identified discussions of medication change (Disposition events), annotators were asked to label the context for the change along four dimensions: Action, Temporality, Certainty, and Actor.

\textbf{Action} indicates the type of change being discussed, and can take one of seven labels:
\begin{itemize}
\addtolength\itemsep{-3mm}
    \item \textbf{Start}: indicates start of a medication patient is not already on
    \item \textbf{Stop}: indicates stop of a medication patient is already on
    \item \textbf{Increase}: indicates an increase in daily dose
    \item \textbf{Decrease}: indicates a decrease in daily dose
    \item \textbf{OtherChange}: indicates a non-dosage related change, such as changes in timing (e.g. take in am instead of pm), change from brand name to generic, or change in formulation (e.g. oral tab to oral solution).
    \item \textbf{UniqueDose}: indicates single administration, where patient has taken a medication but it is unclear from the text whether it's part of a longer planned regimen; often applicable in inpatient or emergency room settings (e.g. ``\emph{In the ED, patient received vancomycin 1 gram}''.
    \item \textbf{Unknown}: used when it is unclear which of the other labels are appropriate to use, for example, ``\emph{Will change to Lasix bid}'' -- unclear if the change is from another medication to Lasix (Start), or if the patient is already on Lasix and is changing to a different frequency (OtherChange) or different dose (Increase or Decrease).
\end{itemize}

\textbf{Temporality} indicates when the change action is intended to occur:
\vspace{-2mm}
\begin{itemize}
\addtolength\itemsep{-3mm}
    \item \textbf{Past}: indicates the action has already taken place
    \item \textbf{Present}: indicates an action intended for the present time
    \item \textbf{Future}: indicates an action that will take place in the future
    \item \textbf{Unknown}: used when it is unclear which of the other labels are appropriate to use
\end{itemize}
\vspace{-2mm}
\textbf{Certainty} indicates whether the change action was implemented or just discussed:
\vspace{-2mm}
\begin{itemize}
\addtolength\itemsep{-3mm}
    \item \textbf{Certain}: indicates definitive action that will take place or has already occurred
    \item \textbf{Hypothetical}: indicates an action being considered but not yet decided upon
    \item \textbf{Conditional}: indicates an action that is dependent upon a specified condition being met
    \item \textbf{Unknown}: used when it is unclear which of the other labels are appropriate to use
\end{itemize}
\vspace{-2mm}
\textbf{Actor} indicates the individual who initiated the change action:
\vspace{-2mm}
\begin{itemize}
\addtolength\itemsep{-3mm}
    \item \textbf{Physician}: indicates a recommendation by the healthcare provider, including physicians, nurse practitioners, or other providers participating in the patient's care
    \item \textbf{Patient}: indicates an action initiated by the patient or their caretaker without consulting their healthcare provider
    \item \textbf{Unknown}: used when it is unclear which of the other labels are appropriate to use
\end{itemize}

Figure \ref{fig:flowchart} presents a flowchart demonstrating how annotators label the Event type (Disposition vs NoDisposition vs Undetermined) for medication mentions, and the Action type for Disposition events.

\section{Experimental Details}\label{apd:second}

\subsection{Feature Classes}
The features used in our experiments can be divided into following classes:
\begin{itemize}
\item\textbf{Lexico-syntactic}: Standard lexical features like n-grams of whole, stemmed and lemmatized words, and part-of-speech tags.

\item\textbf{Window}: Windowed lexico-syntactic features where the positional information with the lexico-syntactic features for tokens that appear within of a window of 5 from the medication in either direction.

\item\textbf{Dependency-parse}. Features which are based on the dependency parse structure of the sentence related the target medication, e.g. neighbors, parent and children of the medication concept in the dependency parse tree.

\item\textbf{RxNorm}. These features were derived from the RxNorm\footnote{\url{https://www.nlm.nih.gov/research/umls/rxnorm/index.html}} for the target medication, such as ATC-class, ingredient, etc.

\item\textbf{Note-section}. We also experimented with rule-based note-section model that classifies each sentence based on the section headers and location of the sentence within the note.

\end{itemize}

\subsection{Experimental Setup}
Experiments were performed using Support Vector Machines (SVM) with one-vs-all classification strategy as implemented in scikit-learn\footnote{\url{https://scikit-learn.org/}}. Initial experiments with different kernels (linear, poly and rbf) showed comparable performance, so we chose linear kernel for all of our experiments.

We used scispaCy\footnote{\url{https://allenai.github.io/scispacy/}}, a biomedical/clinical adapted version of spaCy\footnote{\url{https://spacy.io/}},
for sentence segmentation and extraction of lexico-syntactic and dependency-parse features. As the dataset will be release as a shared task in 2021, we only present our experimental results on the train set. We used 5-fold cross validation throughout our experiments. To deal with skew in class distribution, we experimented with under-sampling and over-sampling but achieved best results by adjusting class weights to be inversely proportional to class frequencies. Finally, we excluded OtherChange for Action dimension and Unknown for all dimensions due to the limited number of instances from our experiments.

\end{document}